%% file: ms.tex
\documentclass[runningheads]{llncs}

\usepackage{graphicx}
\usepackage{amsmath}
\usepackage{amssymb}

\usepackage{comment}

\usepackage{cite}
\usepackage{nicefrac}
\usepackage{flushend}
\usepackage{multirow}
\usepackage{xspace}

\newcommand*{\eg}{\textit{e.g.}\xspace}
\newcommand*{\ie}{\textit{i.e.}\xspace}
\newcommand*{\vs}{\textit{vs.}\xspace}
\newcommand*{\etal}{\textit{et al.}\xspace}
\newcommand*{\etc}{\textit{etc.}\xspace}

\newcommand*{\x}{\mathsf{x}\mskip1mu}

\newcommand{\squishitemize}{
 \begin{list}{$\bullet$}
  { \setlength{\itemsep}{0pt}
     \setlength{\parsep}{3pt}
     \setlength{\topsep}{0pt}
     \setlength{\partopsep}{0pt}
     \setlength{\leftmargin}{1.95em}
     \setlength{\labelwidth}{1.5em}
     \setlength{\labelsep}{0.5em} } }

\newcounter{Lcount}
\newcommand{\squishlist}{
    \begin{list}{\arabic{Lcount}. }
   { \usecounter{Lcount}
        \setlength{\itemsep}{0pt}
        \setlength{\parsep}{3pt}
        \setlength{\topsep}{0pt}
        \setlength{\partopsep}{0pt}
        \setlength{\leftmargin}{2em}
        \setlength{\labelwidth}{1.5em}
        \setlength{\labelsep}{0.5em} } }

\newcommand{\squishend}{\end{list}}

\usepackage[capitalise,noabbrev]{cleveref}
\crefformat{section}{\S#2#1#3} 
\crefformat{subsection}{\S#2#1#3}
\crefformat{subsubsection}{\S#2#1#3}

\begin{document}

\pagestyle{headings}
\mainmatter

\title{Reducing Inference Latency with Concurrent Architectures for Image Recognition} 

\titlerunning{Reducing Inference Latency with Concurrent Architectures} 
\authorrunning{Hadidi \& Cao et al.} 
\author{Ramyad Hadidi$^\dagger$$^1$, Jiashen Cao$^\dagger$$^1$, Michael S. Ryoo$^2$, Hyesoon Kim$^1$}
\institute{Georgia Institute of Technology$^1$, Stony Brooks University$^2$
          \\[2pt]
          \scriptsize{$\dagger$ Same Contribution}}


\maketitle

\begin{abstract}
\input{tex/abstract.tex}
\end{abstract}

\section{Introduction \& Motivation}
\label{sec:intro}
\input{tex/intro.tex}

\section{Related Work}
\label{sec:related}
\input{tex/related}

\section{Concurrent Architectures}
\label{sec:parallelNets}
\input{tex/parallel}

\section{Experimental Analysis}
\label{sec:experiments}
\input{tex/experiments}

\section{Conclusion}
\label{sec:conclusion}
\input{tex/conclusion}



\bibliographystyle{splncs04}
\bibliography{ms}

\newpage
\section{Appendix}
\label{sec:appendix}
\input{tex/appendix}

\end{document}

%% file: tex/abstract.tex
Satisfying the high computation demand of modern deep learning architectures is challenging for achieving low inference latency. The current approaches in decreasing latency only increase parallelism within a layer. This is because architectures typically capture a single-chain dependency pattern that prevents efficient distribution with a higher concurrency (\ie, simultaneous execution of one inference among devices). Such single-chain dependencies are so widespread that even implicitly biases recent neural architecture search (NAS) studies. In this visionary paper, we draw attention to an entirely new space of NAS that relaxes the single-chain dependency to provide higher concurrency and distribution opportunities. To quantitatively compare these architectures, we propose a score that encapsulates crucial metrics such as communication, concurrency, and load balancing. Additionally, we propose a new generator and transformation block that consistently deliver superior architectures compared to current state-of-the-art methods. Finally, our preliminary results show that these new architectures reduce the inference latency and deserve more attention.

%% file: tex/intro.tex
\begin{figure}[t]
    \vspace{-0pt}
    \centering
    \includegraphics[width=0.70\linewidth]{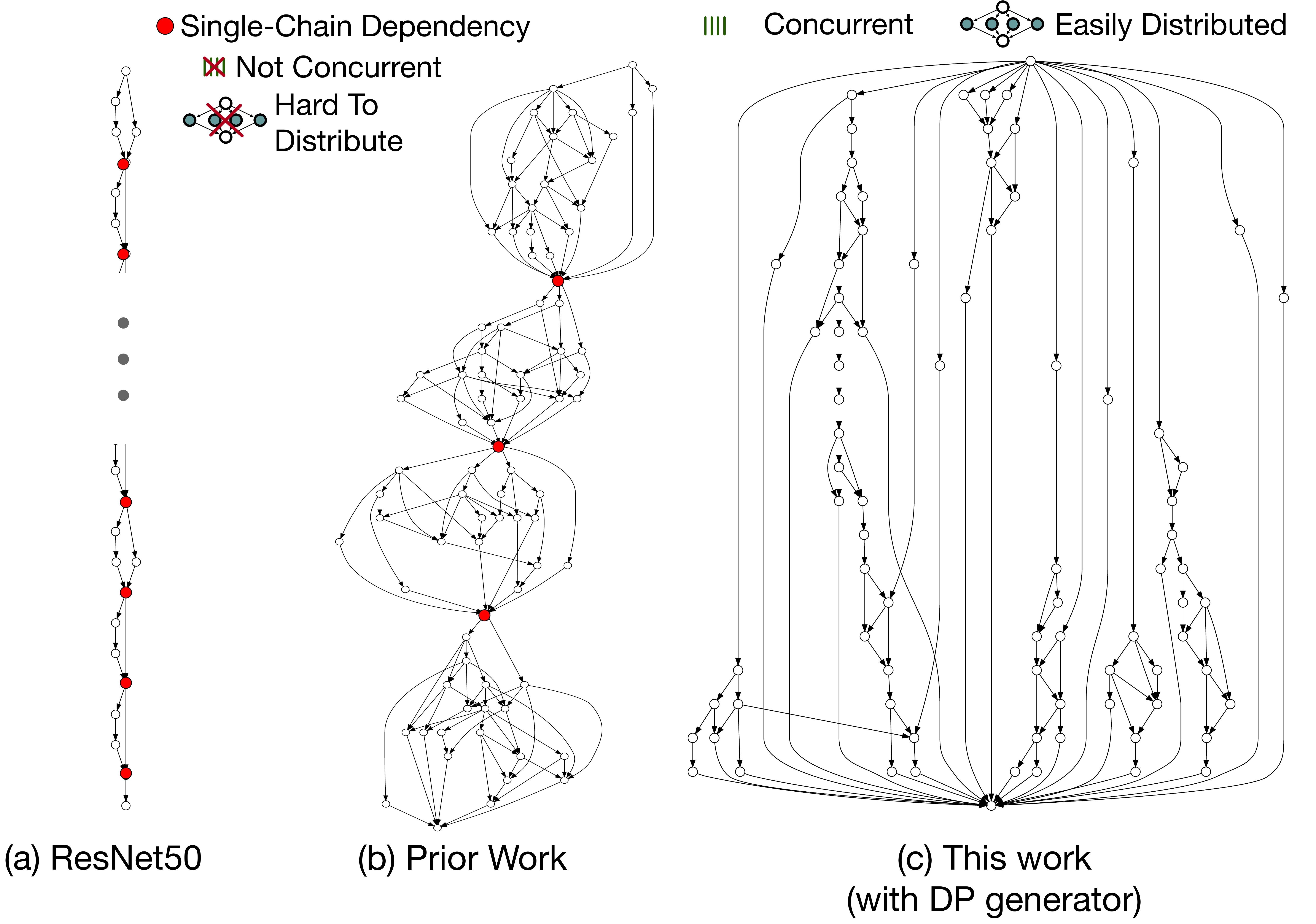}
    \vspace{-8pt}
    \caption{\textbf{Sampled Architectures Overview} -- (a) \& (b) Limited concurrency and distribution due to single-chain dependency. (c) Improved concurrent architecture.}
    \label{fig:intro}
    \vspace{-18pt}
\end{figure}

%
%
%
Increasingly deeper and wider convolution/deep neural networks (CNN/DNN)~\cite{szegedy2017inception, nasnet1, real2019regularized} with higher computation demands are continuously attaining higher accuracies.  Nevertheless, the high computation and memory demands of these DNNs hinder achieving low inference latency~\cite{hadidi2019characterizing}. Although current platforms exploit parallelism, we discover that, since most architectures capture a \emph{single-chain dependency pattern}~\cite{kri:sut12,sim:zis14-deep,redmon2016you}, shown in Figures~\ref{fig:intro}a \& b, we cannot efficiently extend concurrency and distribution beyond current explicit parallelism exposed within intra-layer computations (\ie, matrix-matrix multiplications) to reduce the latency of an inference. In other words, distribution and concurrency, if any, is implemented at data level~\cite{hazelwood2018applied}, which only increases the throughput. 

The status quo approaches in reducing the inference latency are always applied \emph{after} an architecture is defined (\eg, reducing parameters with weight pruning~\cite{han:mao15} or reducing computation with quantization~\cite{van:sen11}). Additionally, for extremely large architectures, limited model parallelism is applied on final layers (\ie, large fully-connected layers that do not fit in the memory~\cite{hadidi2018musical, had:cao18:2, hadidi2020towards}). However, since model-parallelism methods do not change the architecture, distributing all layers with such methods adds several synchronization/merging points, incurring high communication overheads (Figure~\ref{fig:intro}a \& b). We discover that the single-chain inter-layer dependency pattern, common in all the well-known architectures and even in state-of-the-art neural architecture search (NAS) studies~\cite{xie2019exploring}, prevents the efficient model distribution for reducing inference latency.

This visionary paper addresses the single-chain data dependency in current architecture designs and endeavour to inspire discussion for new concurrent architectures. To do so, first, we analyze architectures generated by recent unbiased NAS studies~\cite{xie2019exploring} and discover that \emph{scaling/staging} blocks implicitly enforce dependencies. Then, we generate new architectures with prior and our new distance-based network generators using our new probabilistic scaling block. Then, for quantitatively comparing generated architectures, we propose a \emph{concurrency score} that encapsulates important metrics such as communication, load balancing, and overlapped computations, by reformulating the problem as a hypergraph partitioning problem~\cite{catalyurek1999hypergraph, lengauer2012combinatorial}. Based on the scores and experiments, our generated architectures have higher concurrency and are more efficient for distribution than current architectures, an example of which is shown in Figure~\ref{fig:intro}c. Additionally, as shown in Figure~\ref{fig:acc_ps}, they provide competitive accuracy while delivering high concurrency, directly proportional to inference latency (Figure~\ref{fig:exp:comp}). Our experiment results (on over 1000 samples) show that our architectures achieve 6--7$\x$ faster inference time. As an added benefit, the current methods in reducing the inference latency can be applied on top of our generated architectures.  The following is our contribution:

\begin{figure}[t]
    \vspace{-0pt}
    \centering
    \includegraphics[width=0.8\linewidth]{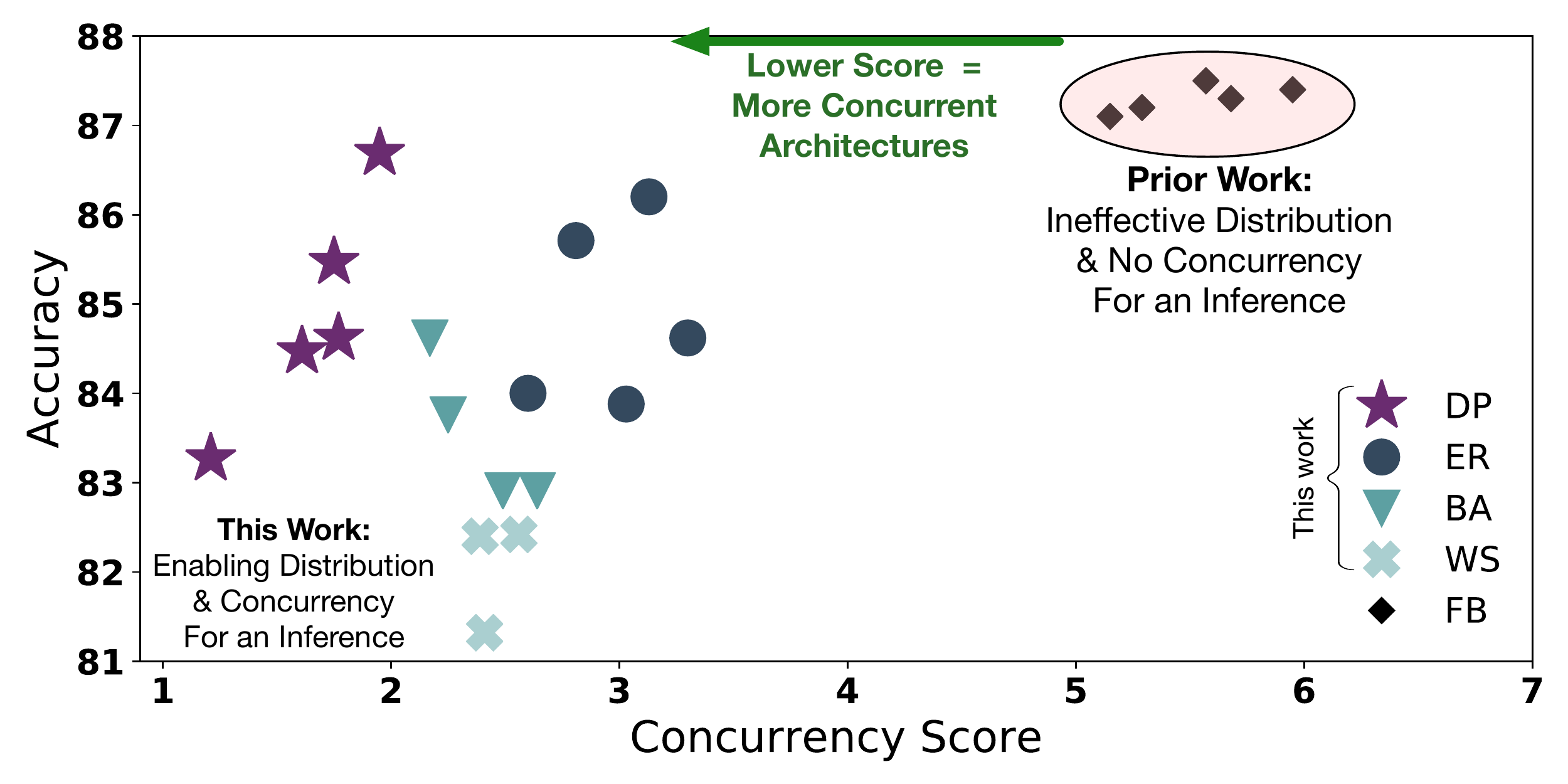}
    \vspace{-10pt}
    \caption{\textbf{Accuracy vs. Concurrency Score} -- Randomly sampled concurrent architectures generated with our NAS consistently achieve competitive accuracies with a higher concurrency and distribution opportunities during an inference (Flower-102, \cref{sec:parallelNets}).}
    \label{fig:acc_ps}
    \vspace{-10pt}
\end{figure}

%
\squishitemize
    \item[] \textbf{Addressing Single-Chain Data Dependencies:} Our concurrent architectures created by network generators (specially the new distance-based generator) break current biased designs by delivering high concurrency.
    \item[] \textbf{Proposing Representative Concurrency Score:} Our problem formulation based on hypergraph theory encapsulates critical metrics to quantitatively compare all architectures for efficient distribution and concurrency.
\squishend
%

%% file: tex/related.tex
\textbf{Computation \& Parameter Reduction:} Reducing computation and parameters to reduce inference latency is an active research area. These techniques are applied after an architecture is fixed. One common approach is to remove the weak connections with weight pruning~\cite{yu:luk17,han:mao15,lin2017runtime, wen:wu16, anwar2017structured}, in which the close-to-zero weights are pruned away.  It is also been shown that moderate pruning with iterative retraining enables superior accuracy~\cite{han:mao15}. Quantization and low-precision inference~\cite{cou:ben14, gon:li14, van:sen11,koster2017flexpoint,lin2016fixed} change the representation of numbers for faster calculations. Several methods also have been proposed for binarizing the weights~\cite{li:zha16, cou:hub:16, rast:ord16}. The concurrent architectures can also benefit from these approaches, making them complementary to further reduce inference latency.

\noindent
\textbf{Concurrency \& Distribution:} With increasingly larger architectures and widespread usage of deep learning, distribution have gained attention~\cite{dean:cor12, mao:chn17, tee:mcd17, kan:hau17, had:cao18:2}. Most of the techniques either exploit data or model parallelism~\cite{kri:sut12, dean:cor12}. \emph{Data parallelism only increases the throughput of the system and does not affect the latency.} Model parallelism divides the work of a single inference. However, model parallelism keeps the connections intact. Thus, applying model parallelism on intra-layer computations results in a huge communication overhead for sharing the partial results after each layer due to existing single-chain dependency. SplitNet~\cite{kim2017splitnet} focuses on improving the concurrency opportunity within an architecture by explicitly enforcing dataset semantics in the distribution of \emph{only} the final layers. Each task needs to be handcrafted individually for each dataset by examining the semantics in the dataset. In this paper, we propose concurrent architectures that is generated by NAS by considering all important factors for distribution, which has not been explored by prior work.

\noindent
\textbf{Neural Architecture Search:} With the growing interests in automating the search space for architectures, several studies~\cite{nasnet1, nasnet2, metaQNN, liu2018progressive, real2019regularized, mnasnet, xie2019exploring} have proposed new optimization methods. Most of these studies~\cite{nasnet1, nasnet2} utilize an LSTM controller for generating the architecture. However, as pointed out in ~\cite{xie2019exploring}, the search space in these studies is determined by the implicit assumption in network generators and sometimes explicit staging (\ie, downsampling spatially while upsampling channels). Although Xie \etal~\cite{xie2019exploring} aimed to remove all the implicit wiring biases from the network generator by using classical random graph generator algorithms, they introduced a scaling/staging bias in the final architecture to deal with a large amount of computation. Such stagings create a merging point after a stage where all the features are collected and downsampled before the next stage. Hence, the generated architecture still carries the single-chain of dependency which limits the further concurrency. In contrast, our proposed architectures do not enforce such a dependency by removing this bias. Moreover, compared to prior work, our target is to reduce inference latency by increasing concurrency, which has not been explored before.

%% file: tex/parallel.tex
Here, we propose concurrent architectures that break the single-chain dependency pattern for enabling concurrent execution of an inference. To improve distribution and concurrency, we aim to search for an architecture that has minimal communication overhead and is load balanced when it is distributed. To do so, the following provides the general problem formulation, while \cref{sec:parallelNets:generators} and \cref{sec:parallelNets:transformations} describe our implementation details. In \cref{sec:parallelNets:parallel}, we extend the representation to quantitatively study distribution and concurrency opportunities, derived by reformulating the problem as a hypergraph partitioning problem.


\noindent
\textbf{Overview:}
The current design of neural architectures is optimized for prediction accuracy and has an implicit bias towards the single-chain approach~\cite{xie2019exploring, nasnet2}, as we discussed in \cref{sec:intro}. This bias limits concurrency and distribution for reducing inference latency. In other words, only the computation within a layer is performed in parallel and not the computation within a model. To tackle this challenge, we aim to consider concurrency and distribution during the design stage and test if such architectures provide higher concurrency with good accuracy. To do so, first, we use network generators to create a random graph structure, which represents a potential architecture. Among all generated architectures, we sample (without any optimized search) and evaluate generated architectures with our proposed concurrency score. Then, we transform the graph to a DNN and perform experiments. Our final results show a promising direction worth exploring. 

\noindent
\textbf{DAG Representation:}
A neural architecture, $\mathcal{N}$, can be represented as a directed acyclic graph (DAG) because the computation flow always goes in one direction without looping. We define a DAG as $\mathcal{G}=(V,E)$ where $V$ and $E$ are sets of vertices and edges, respectively. We define a network generator, $\mathit{f}$, as a function that constructs random DAG. $\mathit{f}$ creates the edge set, $E$, and defines the source and sink vertices for each edge, regardless of the type of the vertices. Although network generators could be deterministic (\eg, a generator implemented with NAS approach), we are interested in stochastic network generators. The reasons are two-fold. First, the stochastic generator provides a larger search space than the deterministic generator, so it is more likely to remove any bias. Second, since, unlike prior work, we don not use scaling/staging to glue different parts of our NAS generated network~\cite{xie2019exploring} (shown in Figure~\ref{fig:intro}b), stochastic generators provide more options for a potential solution. Note that the generated DAG only represents the dataflow and does not include the weights, which are learned in subsequent steps. \cref{sec:parallelNets:generators} provides more details about our network generators and how we utilize them to create a DAG.

\noindent
\textbf{DAG to DNN:} 
Once we have found a promising DAG representation after the concurrency score study, we transform the DAG into an actual DNN. Vertices in DAG are components (\eg, layers or sub-networks) and edges are connections. Within the process of transformation, we convert the nodes in DAG to a block of layers and connect blocks with its corresponding edge in DAG. Each vertex, $\mathit{V}_i$, has several properties such as type of the layer and its properties (\eg, depth, width, activation size, \etc). In this paper, we use a uniform computation in vertices: ReLU, $3\x3$ separable convolution~\cite{chollet16}, and batch normalization~\cite{iof:sze15}.

\subsection{Network Generators}
\label{sec:parallelNets:generators}

We use three classical random graph generators as baselines. Additionally, after discovering that state-of-the-art generators do not generate a concurrent architecture, we propose a new graph generator with distance-based heuristics. Below, we describe the generators identified by how their stochastic nature influences the graph. Note that although the first three generators are based on~\cite{xie2019exploring}, to generate concurrent architectures, we have removed the introduced staging blocks, which enforces the single-chain dependency in prior work. Thus, all the studied architectures in this work are novel and have never been studied before.

Once we obtain an undirected random graph from the generator, we convert the undirected graph to DAG by using the depth-first search algorithm. The vertices with smaller vertex ID is traversed earlier than vertices with larger ID. As the final step, we add an input vertex to all vertices without predecessors and an output vertex to all vertices without successors. This ensures that we obtain a DAG with a single source and sink.



\begin{figure}[b]
    \vspace{-15pt}
    \centering
    \includegraphics[width=0.70\linewidth]{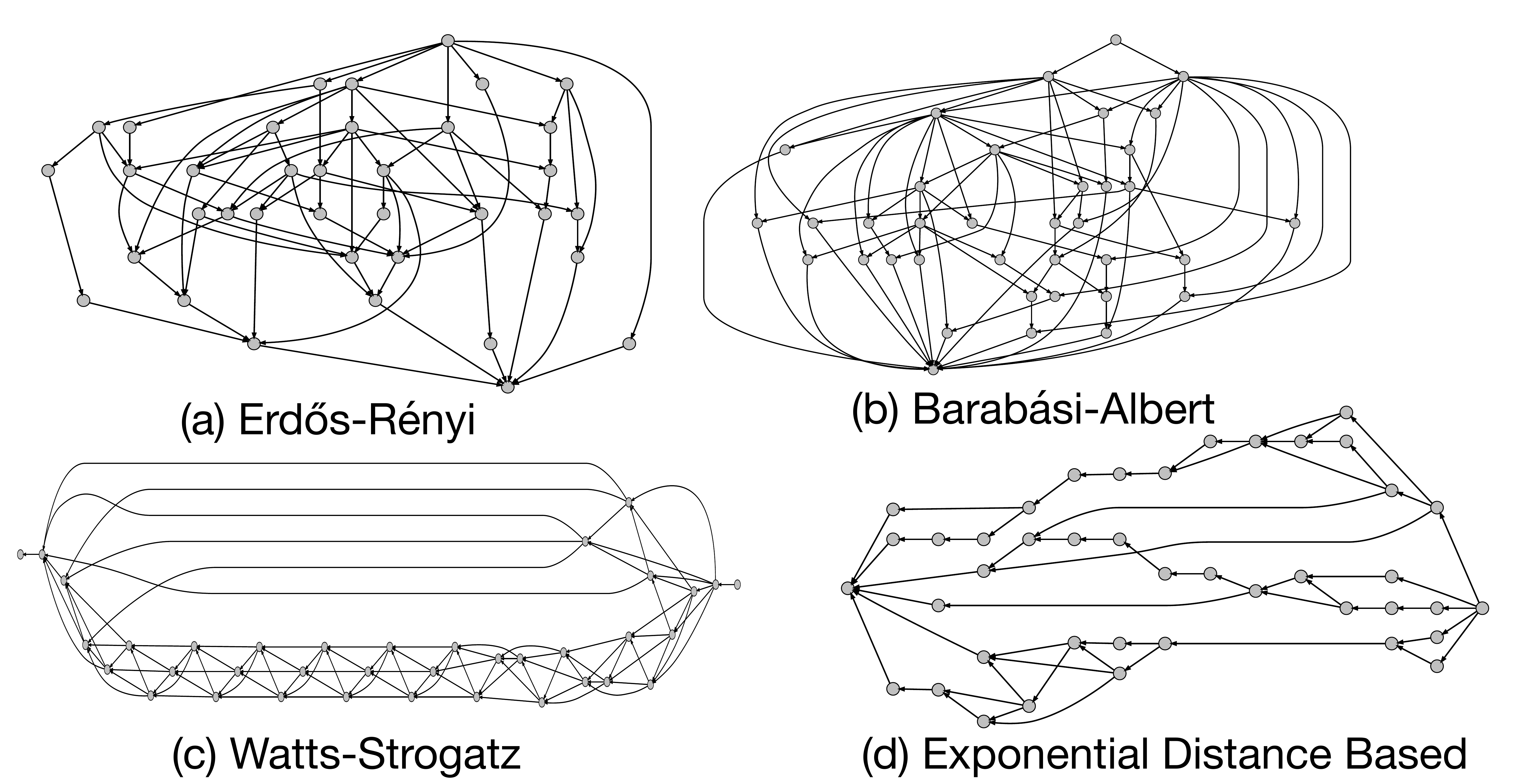}
    \vspace{-10pt}
    \caption{\textbf{Network Generators} -- Four examples of different random graph generators. Note that only (d) produces a good concurrent balanced graph.}
    \vspace{-0pt}
    \label{fig:trans:gen}
\end{figure}

\noindent
\textbf{(1) Independent Probability:} In this group, the probability of adding an edge is independent of other properties. Similar to the Erd{\H{o}}s and R{\'e}nyi model (ER)~\cite{erdHos1960evolution}, in which an edge exists with a probability of $P$. Generators with independent probability completely ignore the graph structure and create a connected graph (Figure~\ref{fig:trans:gen}a) that is hard to efficiently distribute.

\noindent
\textbf{(2) Degree Probability:} In this group, the probability of adding an edge is defined by the degree of one of its connected vertices. A vertex with a higher degree has more probability of accepting a new edge. Figure~\ref{fig:trans:gen}b shows an example of such a generator. Barab{\'a}si-Albert model (BA)~\cite{albert2002statistical}, first adds $M$ disconnected vertices, then for the total number of vertices until $N$, it adds a total of $M$ edges with a linear probability proportional to the degree of each vertex (\ie, a total of $M(N-M)$ edges). Generators with degree probability create a tree-structured graph, in which at least one vertex is strongly connected to other vertices. Such a graph structure is hard to distribute since all the vertices are dependent on at least one vertex, if not more.

\noindent
\textbf{(3) Enforced Grouping:} In this group, initially, a pre-defined grouping is performed on disconnected vertices and then edges are added based on the groups. Small world graphs~\cite{kleinberg1999small, watts1999networks, newman1999renormalization} are good examples. In one approach (WS)~\cite{watts1999networks}, vertices are placed in a ring and each one is connected to $\nicefrac{K}{2}$ neighbors on both sides. Then, in a clockwise loop on vertices, an existing edge between its $i_{th}$ neighbor is rewired with a uniform probability of $P$ for $\nicefrac{K}{2}$ times. As shown in Figure~\ref{fig:trans:gen}c, a graph with WS algorithm tends to form a single-chain structure if $P$ is small. With a larger $P$, the structure becomes similar to ER.

\noindent
\textbf{(4) Distance Probability:} In distance probability (DP), initially, a pre-defined grouping is performed on disconnected vertices, then a distance probability function defines the existence of an edge. We first arrange the vertices in a ring. Then, the probability of adding an edge between two vertices is dependent on their distance. In other words, closer vertices have a higher probability of getting edges.

\noindent
\emph{$-$ Distance Metrics:} We define distance $d$ as the smallest number of nodes plus one between two nodes in a ring. The maximum distance can be half of the total number of nodes, which is $\nicefrac{N}{2}$. We use the distance to re-scale the passed in probability $\mathit{P}$ presented in WS. We use exponential re-scaling function:
\begin{equation}
\small
    P_\text{new} = \alpha P^{\beta d},
\end{equation}
in which $\alpha$ and $\beta$ are constants. The probability quickly decreases as the distance increases. This mechanism naturally creates multiple locally strongly connected graphs, Figure~\ref{fig:trans:gen}d, which can be distributed. However, we still need to examine the distribution and concurrency opportunities, which are presented in \cref{sec:parallelNets:parallel}.


\subsection{Transformations}
\label{sec:parallelNets:transformations}

Transformations are operations, the main objective of which is to create a reasonable architecture, that happens after the construction of the DAG. We first introduce the building blocks, which include a scaling building block that, contrary to previous work, does not enforce a single-chain dependency. 


\begin{figure}[t]
    \vspace{-0pt}
    \centering
    \includegraphics[width=0.50\linewidth]{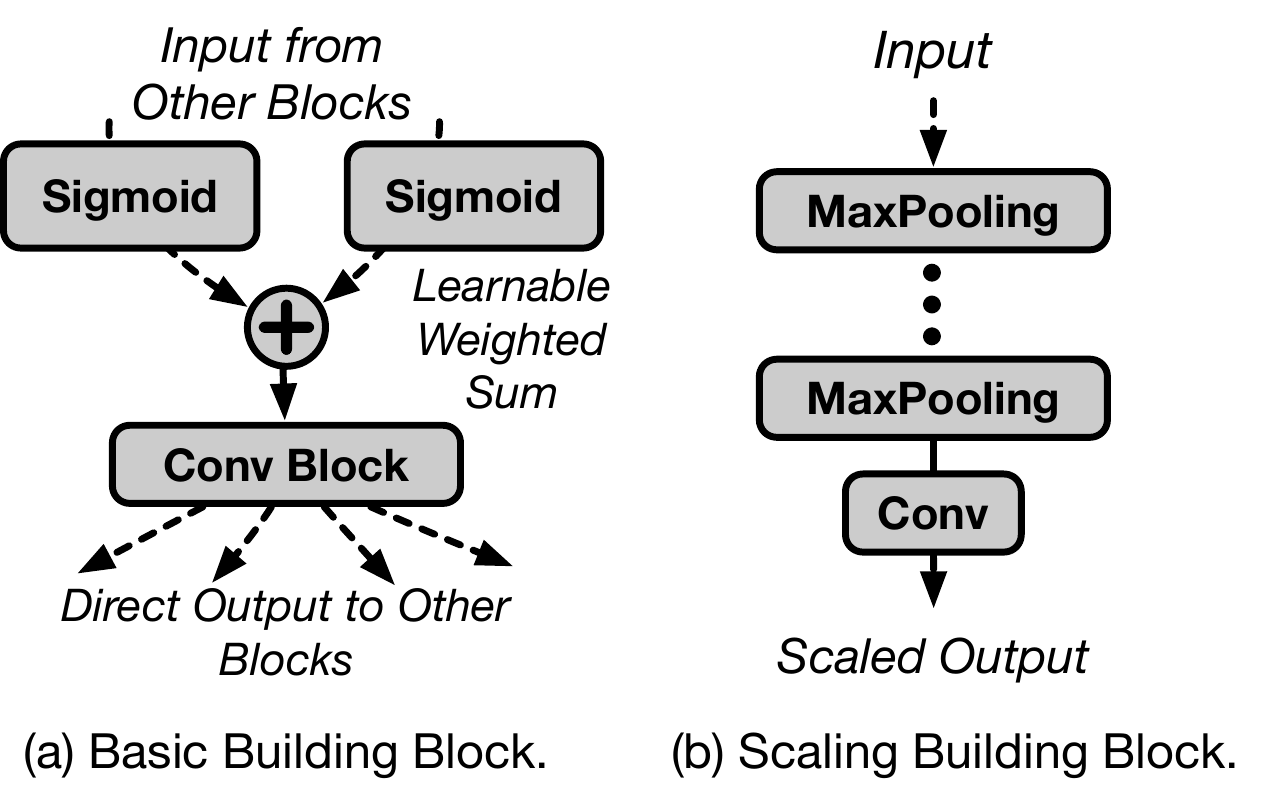}
    \vspace{-10pt}
    \caption{\textbf{Building Blocks} -- Building blocks used for conversion from DAG to DNN.}
    \label{fig:trans:blk}
    \vspace{-2pt}
\end{figure}

\noindent
\textbf{Building Block:} During the process of transforming a DAG to DNN, vertices are interpreted as basic building blocks, as shown in Figure~\ref{fig:trans:blk}. Inside a basic building block, Sigmoid activations are applied on inputs, then, the activations are summed with a learnable weighted sum. The Sigmoid function is used to avoid weighted sum overflow. As described before, the \texttt{conv} block consists of a ReLU, 3$\times$3 separable convolution, and batch normalization.

\noindent
\textbf{Redefining Staging:} Staging is deemed to be necessary for all NAS generated architectures to reduce the computation and facilitate learning. For staging, after a few layers, usually, the common method is to gather and merge outputs from all transformation vertices, conduct downsampling, and channel upsampling. However, such staging points create a rigid architecture with single-chain dependencies that are hard to distribute and execute concurrently (\eg,~\cite{xie2019exploring}). To address the single-chain bottleneck problem caused by staging, the first solution is implementing a uniform channel size for the entire architecture. In other words, all \texttt{conv} blocks share the same filter size. Thus, there would be no need to merge and synchronize at a point during an inference. However, as shown in Table~\ref{tb:uniform_channel}, the uniform channel size approach works well on a small image dataset (\eg, Cifar-10), but it fails to achieve good accuracy on a dataset with larger image dimension (\eg, Flower-102). 

\input{tables/uniform_channel}

In this paper, we propose individual staging after any \texttt{conv} block. Because of that, inputs to a \texttt{conv} block could have different dimensions. To tackle this problem, we dynamically add a new scaling block in the process of construction. The scaling block consists of a number of maxpooling layers. Maxpooling layers downsamples the dimensions to match with the smallest dimension in the input. We also use 1$\times$1 convolution layers to upsample the channel size to match the highest channel size in the inputs in these scaling blocks. Therefore, we avoid bottlenecks in generated architecture.

We adopted two design choices for the staging mechanism. In the first design, greedy-based staging, we start with greedy-based staging. Within the construction process, we set an upper limit for channel size. As long as channel sizes have not reached the upper bound, we conduct staging (\ie, downsample the input \& upsample the channel). However, this design raises an issue that intermediate outputs are quickly squeezed through the maxpooling layer, which discards important features. This approach hurts the accuracy to some extent. s
In the second design, probabilistic-based staging, we use a probabilistic method in staging. In this design, although the channel size may have not reached the limit, staging is done with a fixed probability of 0.5 to avoid discarding features too quickly. As shown in Tables~\ref{tb:greedy_vs_prob_acc} and~\ref{tb:greedy_vs_prob_acc_to_param}, the probabilistic approach achieves better accuracy rate than the greedy-based approach. In addition, Table~\ref{tb:greedy_vs_prob_acc_to_param} shows that probabilistic staging supports higher accuracy with less parameter size because (i) probabilistic staging gracefully discards features, so the architecture learns better; and (ii) the aggressive greedy-based staging creates more size mismatch, so it requires more scaling blocks. 

\input{tables/greedy_vs_prob_acc}
\input{tables/greedy_vs_prob_acc_to_param}

\subsection{Concurrency \& Distribution}
\label{sec:parallelNets:parallel}

Our goal in this paper is to inspire concurrent architecture designs to improve inference latency performance. As a result, besides common accuracy consideration, we need to study concurrency and distribution opportunities of a candidate architecture. To help the community to extend our study, instead of focusing and showcasing on a single architecture, we are interested in finding a customized \emph{concurrency score} ($\textsc{CS}$) for a given architecture, $\mathcal{N}$, that is easily calculated. In this way, we can study various architectures and future works that can further improve this work. $\textsc{CS}$ shows how optimal the concurrent and distributed task assignment for an architecture is. Lower $\textsc{PS}$ score represents fewer communications, better load-balanced tasks, and more distribution opportunities with more overlapped computation, so the architecture is more efficient for concurrency.

\noindent
\textbf{Metrics in The Score:} 
We can formulate our problem of allocating tasks on $n$ units as a multi-constraint problem. The first constraint is that all units should perform the same amount of work, or be load balanced. Second, the communication amount, the main bottleneck in distribution, should be at a minimum. And third, we want to minimize runtime by increasing overlapped computations among the units. 
The first two constraints are addressable by finding a set of hypergraph partitions, in which we divide the vertices into equally weighted sets so that few hyper-edges cross between partitions. The derivable metric is the amount of variability in loads ($\delta_W$) and a total of communication ($\Lambda$). 
The third constraint is measurable by finding the longest path between the input and output vertices on the DAG and quantify concurrency ($\eta$). For instance in pipeline parallelism, the longest path is the entire architecture, as a result the latency is never reduced (and throughput is increased). Now, we provide the formal definition of these solutions by first studying the DAG.

\begin{figure}[t]
    \vspace{-0pt}
    \centering
    \includegraphics[width=0.56\linewidth]{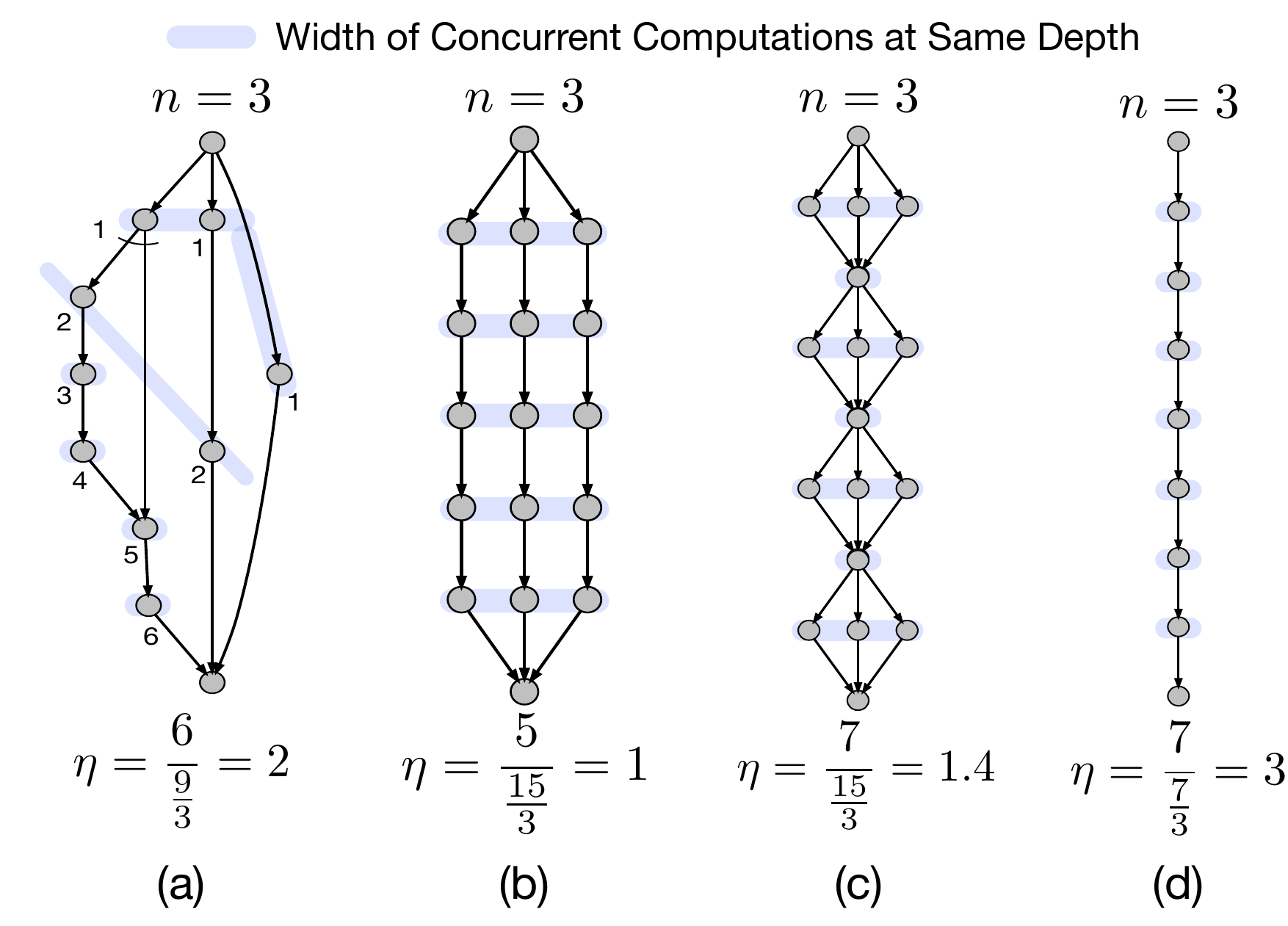}
    \vspace{-5pt}
    \caption{\textbf{Overlapped of Computation Metric} -- Illustration of $\eta$.}
    \label{fig:concurrency}
    \vspace{-15pt}
\end{figure}

\noindent
\textbf{Maximizing Overlapped Computations:}
We measure how overlapped is the inter-layer computations of an architecture from its DAG, or $\eta$, as a raito. We measure this by observing the longest path in the distinct paths between input and output vertices in the DAG, $\mathcal{G}$, relative to the number of the computation cores, $n$. Assume $\{d_i\}$ is the set of distinct longest paths in $\mathcal{G}$. We define $\eta$ as
\begin{equation}
    \small
    \eta = \frac{\text{max}\{d_i\}}{\nicefrac{|\mathcal{V}|}{n}},
\end{equation}
in which $|\mathcal{V}|$ is the total number of vertices.  Figure~\ref{fig:concurrency} depicts an examples of $\eta$. A higher $\eta$ value shows a more limited opportunity to overlap the computation. Figure~\ref{fig:concurrency} also shows the width ofthe overlapped computation at the same depth (\ie, DFS depth with the source of input), which is a good representation of why some architectures are more efficient for concurrency.

\noindent
\textbf{Hypergraph Representation:} 
Using graph representations in task assignment for distributed computing is a well-known problem~\cite{hendrickson2000graph}. Basically, in the generated DAG, vertices of the graph represent the units of computations, and edges encode data dependencies. We can indicate the amount of work and/or data, by associating weights ($w$) and costs ($\lambda$) to vertices and edges, respectively. However, a DAG representation does not sufficiently capture the communication overhead, load balancing factor, and the fact that some edges are basically sending the same data/features. Therefore, for task assignment, we use an alternative graph representation, derivable from the DAG, hypergraph. A hypergraph~\cite{catalyurek1999hypergraph} is a generalization of a graph, in which an edge can join any number of vertices~\cite{hypergraph}. The hypergraph representation, common in optimization for integrated circuits~\cite{lengauer2012combinatorial}, enables us to consider the mentioned factors.
%

\noindent
\emph{Formal Definition of Hypergraph}: A hypergraph $\mathcal{H} = (\mathcal{V}, \mathcal{E})$ is defined as a set of vertices $\mathcal{V}$ and a set of hyper-edges $\mathcal{E}$ selected among those vertices. Every hyper-edge $e_j \in \mathcal{E}$ is a subset of vertices, or $e_j \subseteq \mathcal{V}$. The size of a hyper-edge is equal to the number of vertices.

\noindent
\textbf{Hypergraph Partitioning:} 
We assign weights ($w_i$) and costs ($\lambda_j$) to the vertices ($v_i \in \mathcal{V}$) and edges ($e_j \in \mathcal{E}$) of the hypergraph, respectively. $\mathcal{P} = \{V_1, V_2, V_3, ..., V_P\}$ is a P-way partition of $\mathcal{H}$ if (i) $\forall V_i, \emptyset \neq V_i \subset \mathcal{V}$, (ii) parts are pairwise disjoint, and (iii) $\bigcup \mathcal{P} = \mathcal{V}$. A partition is balanced if $W_p \leq \varepsilon W_\text{avg} $ for $1 \leq p \leq P$, where $W_\text{avg}= \nicefrac{\sum_{v_i \in \mathcal{V}} w_{v_i}}{P}$ denotes the weight of each part, and $\varepsilon$ represents the imbalance ratio, or $\delta_W$.

In a partition $\mathcal{P}$ of $\mathcal{H}$, a hyper-edge that has at least one vertex in a part is said to connect that part. The number of connections $\gamma_j$ of a hyper-edge $e_j$ denotes the number of parts connected by $e_j$. A hyper-edge is a cut if $\gamma_j > 1$. We define such hyper-edges as an external hyper-edges $\mathcal{E}_E$. The total communication for $\mathcal{P}$ is
\begin{equation}
    \small
    \Lambda = \sum_{e_j \in \mathcal{E}_E} \lambda_j(\gamma_j-1).
\end{equation}
Therefore, our two constraints can be defined as a hypergraph partitioning problem, in which we divide a hypergrpah into two or more parts such that the total communication is minimized, while a given balance criterion among the part weights is maintained. We can solve this NP-hard~\cite{lengauer2012combinatorial} problem with multi-paradigm algorithms, such as hMETIS~\cite{karypis1999multilevel} relatively fast. Note that solving this problem is a pre-processing step, which does not affect runtime. 

\begin{figure}[t]
  \vspace{0pt}
  \centering
  \includegraphics[width=0.6\linewidth]{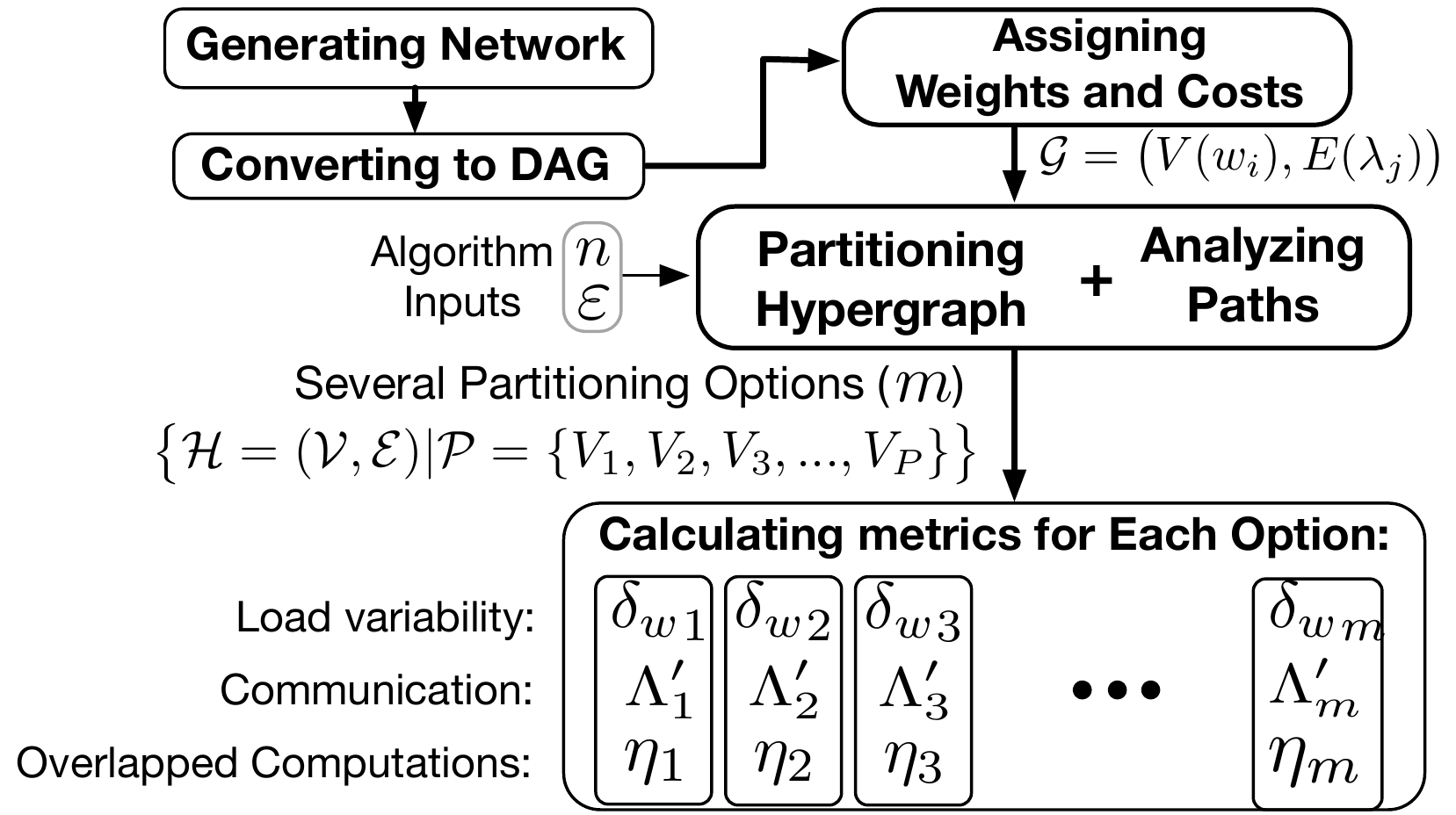}
  \vspace{-10pt}
  \caption{\textbf{Calculating Concurrency Score} -- Summarizing steps for deriving the score.}
  \vspace{-15pt}
  \label{fig:algo}
\end{figure}

\noindent
\textbf{Concurrency Score:} 
Now, we have the tools to calculate the concurrency score, $\textsc{CS}$. Figure~\ref{fig:algo} summarizes all the steps to derive our metrics: Load variability, $\delta_w$; total amount of communication, $\Lambda$; and overlapped computations, $\eta$. Hypergraph algorithm accepts the number of units and a higher bound of $\varepsilon$. By changing the $\varepsilon$, we create a set of partitioning options, for each of which we compute all the metrics. Note that the DAG input requires a weight and cost value for every vertex and edge, respectively. Both of these values are easily derivable. The weight of a vertex is directly proportional to its floating operations (FLOPs), reported by most frameworks. The cost of an edge is directly proportional to the transferred data size. 
To get $\textsc{CS}$, first, we need to normalize the communication metric. We write $\Lambda$ as $\Lambda'=\nicefrac{\Lambda}{(U_c\times n)}$, in which $U_c$ is a unit of data and $n$ is the number of units. We define
\begin{equation}
    \small
    \textsc{CS} = \sqrt[\nicefrac{1}{3}]{\delta_w^a \Lambda'^b \eta^c},
    \label{equ:cs}
\end{equation}
as a custom concurrency score, in which $a,b$ and $c$ are constant that show the relative importance of each metric for a user. In this paper, we assume $a=c=1$ and $b=1.5$, for a higher priority for communication. We chose $U_c$ as the smallest amount of communication for an edge in a generator. Hence, a higher $\textsc{CS}$ value shows poor distribution and concurrency opportunities.

%% file: tables/uniform_channel.tex
\renewcommand{\arraystretch}{1.0}
\begin{table}[t]
  \centering
  \footnotesize
  \begin{tabular}{lcc}
    \hline
    
    \multirow{2}{0.4\linewidth}{\textbf{Dataset}}
    & \multirow{2}{0.2\linewidth}{\centering\textbf{Baseline}}
    & \multirow{2}{0.4\linewidth}{\centering\textbf{DNNs with Uniform Channels}} \\ \\
    
    \hline
    
    \textbf{Cifar-10 32$\times$32}
    &  80.70
    &  81.13 \\
    
    \textbf{Flower-102 224$\times$224}
    & 87.80
    & 74.73  \textbf{\footnotesize(Fails to Scale!)} \\
    
    \hline
  \end{tabular}%
  \vspace{-0pt}
\caption{\textbf{Accuracy of Uniform Channels} -- The mean accuracy comparison between sampled group architectures with uniform channel \vs handcrafted without any advanced optimizations. (baselines Cifar-10 and Flower-102 are vanilla CifarNet and ResNet-50, respectively).}
\label{tb:uniform_channel}
\vspace{-25pt}
\end{table}
\renewcommand{\arraystretch}{1.0}

%% file: tables/greedy_vs_prob_acc.tex
\renewcommand{\arraystretch}{1.0}
\begin{table}[t]
  \vspace{-0pt}
  \centering
  \small
  \begin{tabular}{lcccc}
  
    \hline
    \footnotesize{Staging/Samples}
    & \textbf{A}
    & \textbf{B}
    & \textbf{C}
    & \textbf{Overall Mean} \\
    
    \hline
    
    \textbf{Greedy}
    & 82.30~~
    & 81.32~~
    & 82.42~~
    & 82.01~~ \\
    
    \textbf{Probabilistic}
    & 82.42~~
    & 86.69~~
    & 84.62~~
    & \textbf{84.58~~} \\

  \end{tabular}%
  \vspace{0pt}
\caption{\textbf{Average Accuracy} -- Comparison of randomly sampled group of generated architectures with different staging choices (trained on Flower-102).}
\label{tb:greedy_vs_prob_acc}
\vspace{-10pt}
\end{table}
\renewcommand{\arraystretch}{1.0}

%% file: tables/greedy_vs_prob_acc_to_param.tex
\renewcommand{\arraystretch}{1.0}
\begin{table}[t]
  \vspace{-5pt}
  \centering
  \small
  \begin{tabular}{lcccc}
  
    \hline
    \footnotesize{Staging/Samples}
    & \textbf{A}
    & \textbf{B}
    & \textbf{C}
    & \textbf{Overall Mean} \\
    
    \hline
    
    \textbf{Greedy}
    & 2.31~~
    & 2.27~~
    & 2.63~~
    & 2.40~~ \\
    
    \textbf{Probabilistic}
    & 3.00~~ 
    & 3.28~~
    & 3.58~~
    & \textbf{3.29~~} \\

  \end{tabular}%
  \vspace{0pt}
\caption{\textbf{Average Accuracy/Parameters Ratio} -- Comparison of randomly sampled generated architectures with different staging choices (trained Flower-102).}
\label{tb:greedy_vs_prob_acc_to_param}
\vspace{-20pt}
\end{table}
\renewcommand{\arraystretch}{1.0}

%% file: tex/experiments.tex
In this section, we evaluate our generated architectures by comparing our customized generator and transformation process with prior work. The results demonstrate that our generated architectures preserves accuracy while achieving better concurrency scores by removing the implicit bias of single-chain dependency. Besides, by running the final architecture on actual devices, we show that the concurrency score provides reasonable heuristic about the real performance.

\subsection{Experimental Setup}
\label{sec:experiments:imp}

\noindent
\textbf{Generators:} All generators use probabilistic scaling blocks. FB represents prior work in unbiased NAS with staging blocks~\cite{xie2019exploring}. As mentioned in~\cref{sec:parallelNets:generators}, although ER, BA, and WS generators are based on~\cite{xie2019exploring}, we remove the staging block that causes the limited concurrency. As a result, all the studied network generators and resulted architectures are novel and have never been studied before. 

\noindent
\textbf{Randomization:} To evaluate the accuracy of randomly generated architecture, we collect representative samples with \textit{no optimized search}. we followed the same training procedure for architectures and reported the average accuracy. For \textsc{CS}, total communication, and computation time evaluations, we collect 1,000 samples with no optimized search and compare across different generators. 

\noindent
\textbf{Datasets:} We conducted experiments on multiple datasets to ensure the extensibility of concurrent architectures. We use two image classification datasets; (i) Cifar-10~\cite{cifar10}, which contains 60K 32$\times$32 images in 10 classes; and (ii) Flower-102~\cite{flower-102}, which contains 16K 224$\times$224 images in 102 classes. We strongly encourage future extensive studies on larger datasets, but given the heavy-compute bound of NAS-based experiments, we chose to use representative datasets studied in most of the prior works~\cite{wistuba2019survey}.

\noindent
\textbf{Training Procedure:} We use a uniform training pipeline with a stochastic gradient descent optimizer for all architectures. We train on Cifar-10 with 100 epochs and on Flower-102 with 300 epochs. We report the top-1 classification accuracy on the test sets. For the first 100 epochs, we set the learning rate to be 1e-3 and momentum to be 0.9. We changed the learning rate to 5e-4 and momentum to 0.95 for the remaining 200 epochs on Flower-102.

\noindent
\textbf{Implementation:} We implemented all graph representations in Python NetworkX~\cite{hagberg2008exploring} library. Then, we convert a graph to a PyTorch~\cite{pytorch} compatible model. We constructed a graph-based forwarding path in PyTorch module class to directly reproduce the graph structure. 

\begin{figure}[t]
  \vspace{-0pt}
  \centering
  \includegraphics[width=0.92\linewidth]{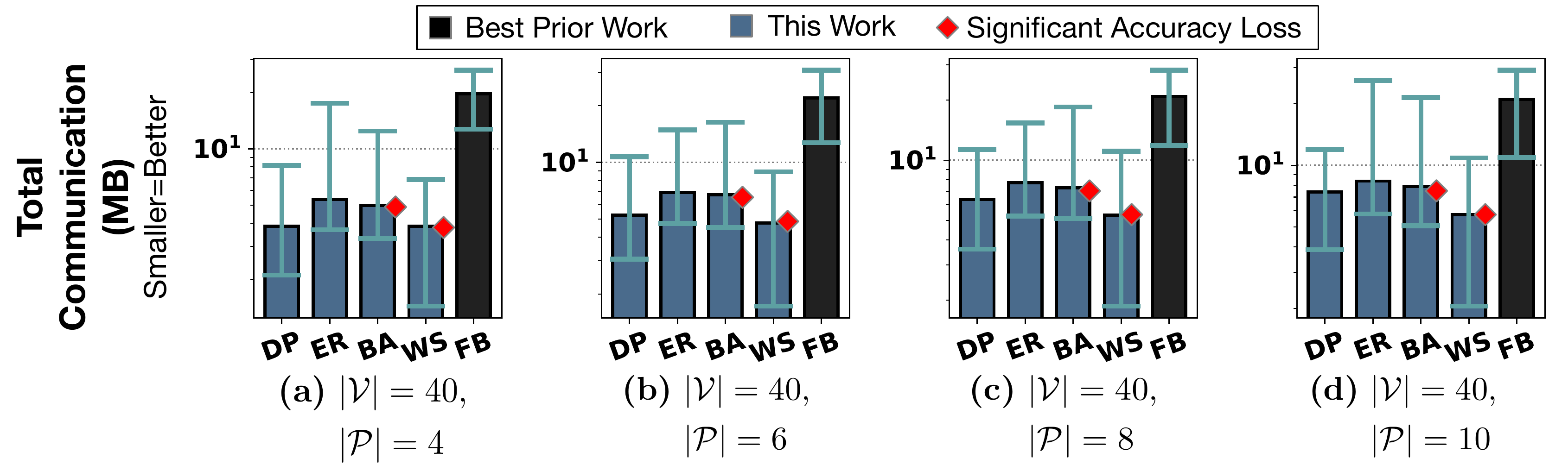}
  \vspace{-10pt}
  \caption{\textbf{Total Communication with Distribution} -- Measured communication in MB for 1000 sampled architectures in each category for 40 vertices on \{4,6,8,10\} units.}
  \vspace{-5pt}
  \label{fig:exp:comm}
\end{figure}

\begin{figure}[t]
  \vspace{-0pt}
  \centering
  \includegraphics[width=0.92\linewidth]{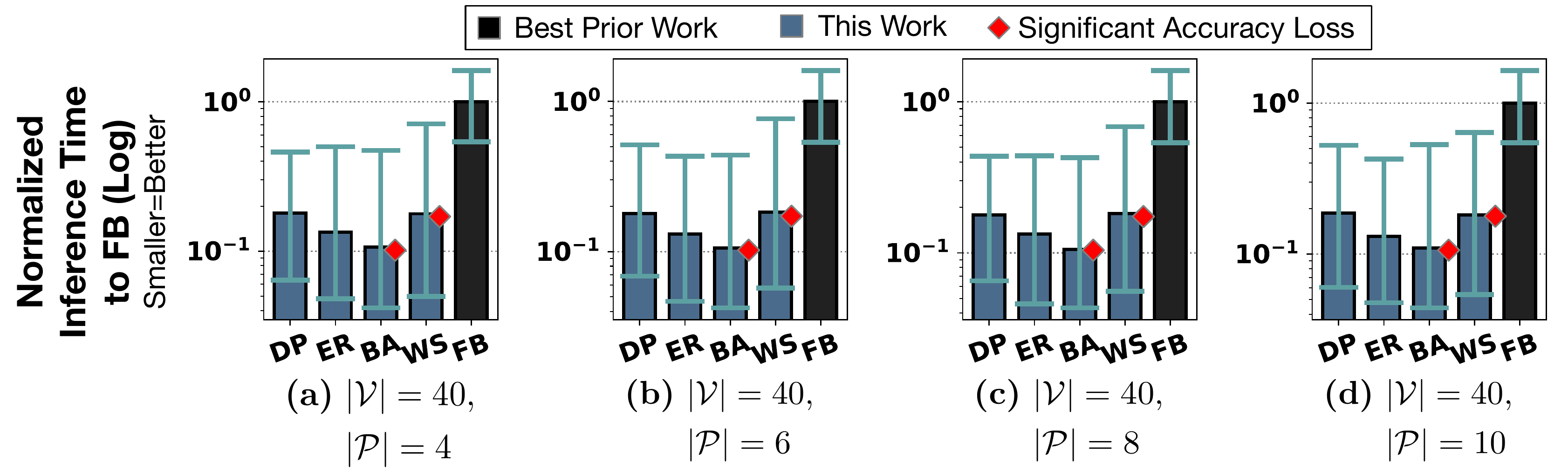}
  \vspace{-10pt}
  \caption{\textbf{Inference Time} -- Normalized inference time normalized to FB (\cref{sec:experiments:imp}) for 1000 sampled architectures in each category for 40 vertices on \{4,6,8,10\} units.}
  \vspace{-15pt}
  \label{fig:exp:comp}
\end{figure}


\begin{figure}[t]
  \vspace{0pt}
  \centering
  \includegraphics[width=0.90\linewidth]{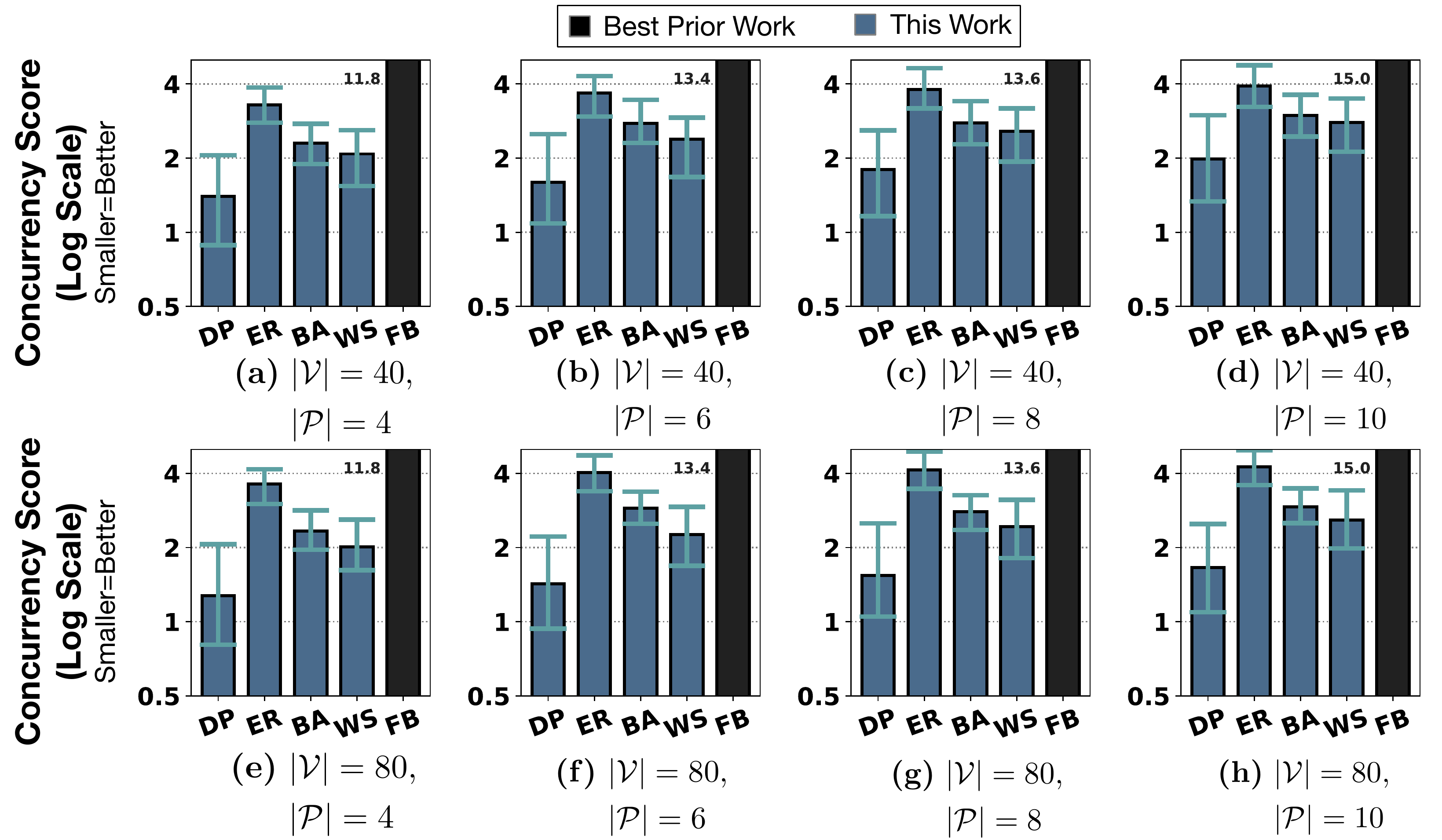}
  \vspace{-10pt}
  \caption{\textbf{Concurrency Scores} -- Measured CS for 1000 sampled architectures in each category with \{40,80\} vertices on \{4,6,8,10\} units (\cref{sec:experiments:imp}).}
  \vspace{-15pt}
  \label{fig:exp:ps}
\end{figure}

\subsection{Experiments}
We analyze the results from three perspectives, communication, latency, and concurrency score. Because we are interested in finding a general solution,  we start with the architecture stability evaluation that particularly focuses on the architecture parameter size. Then, we show the generated architectures achieve competitive accuracies, while, in the last part, we illustrate the high concurrency and distribution opportunities of these architectures.

\noindent
\textbf{Architecture Stability:}
\label{sec:experiments:stability}

For the architecture stability experiment, we used a fixed number of 40 building blocks. We created 1,000 samples from each network generator. We recorded mean and standard deviation regarding the parameter sizes. We also evaluate the architecture stability under different staging design choices (greedy vs probabilistic). From Table~\ref{tb:net_stable_prob}, we see that proposed generators with greedy scaling blocks creates larger but more stable architectures than with probabilistic scaling blocks. Additionally, we see that our proposed DP generator creates the most efficient architecture. We will see that architectures that use DP generators are generally the most optimized. 

\input{tables/net_stable}

\input{tables/cifar}
\input{tables/flower}
%

\noindent
\textbf{Accuracy Study:}

\label{sec:experiments:accuracy}
Here, we demonstrate that the concurrent architectures achieve competitive accuracy on both Cifar-10 and Flower-102 datasets. Given the heavy-compute bound of NAS-based experiments, we encourage further studies on larger datasets. We used the same architecture samples as before without any optimized search and reported both mean and best results. As shown in Table~\ref{tb:cifar} and~\ref{tb:flower}, our concurrent architectures achieve comparable accuracy on both datasets. Generated DNNs achieve better or similar accuracy on Cifar-10. For Flower-102, because both network generation and transformation processes have more randomness, the mean accuracy has a small gap compared to the baseline. However, the best accuracy is close to the baseline, so we believe the accuracy gap can be leveraged by conducting an optimized search in terms of accuracy.

\noindent
\textbf{Concurrency Study:}

\begin{figure}[b]
    \vspace{-15pt}
    \centering
    \includegraphics[width=1.0\linewidth]{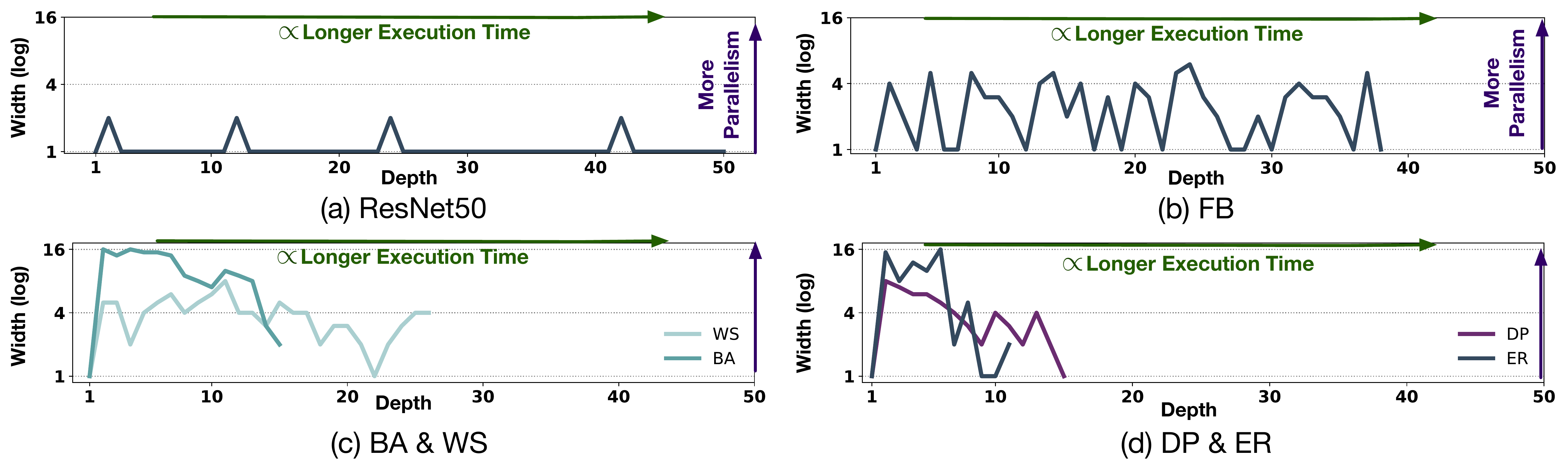}
    \vspace{-18pt}
    \caption{\textbf{Width/Depth Histograms} -- Illustration of ResNet50, FB, and concurrent architectures, which enable more concurrency and shorter inference latency.}
    \label{fig:exp:depth-width}
    \vspace{-0pt}
\end{figure}

\label{sec:experiments:parallel}
Finally, to show improved distribution and concurrency opportunities, we examined the concurrency score of our architectures to ResNet-50 and FB (\cref{sec:experiments:imp}) by sketching width/depth histograms in Figure~\ref{fig:exp:depth-width}. As shown, we achieve higher width/depth, which enables more concurrency, while provides lower maximum depth, which enables shorter execution time. To quantitatively compare the generators and FB, Figure~\ref{fig:exp:ps} depicts concurrency scores, summarized on over 1000 architectures in each category per set. As seen, our generators (and specifically DP) consistently gain the best score. Moreover, to gain more insights, Figure~\ref{fig:exp:comm} and \ref{fig:exp:comp} illustrate total communication with distribution and inference (\ie computation) time, when each architecture is deployed on $|\mathcal{P}|$ units. We see that though ER and BA methods deliver better computation speedup, they suffer performance slow down more from data communication. For our new generator, DP, we see an  6--7$\x$ speedup in inference time. We observe a close relationship between the reported score and actual latency and communication. In fact, latency and communication measure performance in an orthogonal way, but CS score captures the overall efficiency of the generated architecture pretty well and could be used in future studies.

%% file: tables/net_stable.tex
\renewcommand{\arraystretch}{1.0}
\begin{table}[b]
  \vspace{-20pt}   
  \centering
  \small
  \begin{tabular}{p{2.5cm} lcccc}
  
    \cline{3-6}
    
    &
    & \textbf{ER}
    & \textbf{AB}
    & \textbf{WS} 
    & \textbf{DP} \\
    \hline

    Greedy 
    & \textbf{Mean~~~}
    & 48.63~~~
    & 48.33~~~
    & 42.03~~~
    & 35.03~~~\\
    
    Staging
    & \textbf{Std~~~}
    & 1.11~~~
    & 0.91~~~
    & 1.28~~~
    & 2.25~~~ \\

    \\[-10pt]

    \multicolumn{5}{c}{} \\
    \hline
    
    Probabilistic 
    & \textbf{Mean~~~} 
    & 46.03~~~
    & 45.63~~~
    & 36.44~~~
    & 26.69~~~\\
    
    Staging
    & \textbf{Std~~~}
    & 2.70~~~ 
    & 4.41~~~
    & 3.52~~~
    & 3.05~~~\\

  \end{tabular}%
  \vspace{2pt}
\caption{\textbf{Parameter Size Stability} -- The mean and standard deviation of parameter size in sampled generated architectures with different staging.}
\label{tb:net_stable_prob}
\vspace{-0pt}
\end{table}
\renewcommand{\arraystretch}{1.0}

%% file: tables/cifar.tex
\renewcommand{\arraystretch}{1.0}
\begin{table}[t]
  \centering
  \small
  \resizebox{0.75\linewidth}{!}{%
  \begin{tabular}{lcccc}
  
    \hline
    
    \multirow{2}{*}{}
    & \multirow{2}{0.12\columnwidth}{\centering\textbf{Mean Acc.}}
    & \multirow{2}{0.12\columnwidth}{\centering\textbf{Best Acc.}}
    & \multirow{2}{0.20\columnwidth}{\centering\textbf{Mean Acc./Param.}}
    & \multirow{2}{0.20\columnwidth}{\centering\textbf{Best Acc./Param.}} \\ \\
    
    \hline
    
    \textbf{CifarNet}
    & 80.70
    & 80.70
    & 5.38
    & 5.38 \\
    
    \textbf{ER}
    & 81.33
    & 81.81
    & 4.94
    & 5.03 \\
    
    \textbf{BA}
    & 80.29
    & 81.66
    & 4.81
    & 4.92 \\
    
    \textbf{WS}
    & 79.89
    & 81.45
    & 4.75
    & 4.84 \\
    
    \textbf{DP}
    & 80.87
    & 82.47
    & 4.81
    & 4.90 \\

  \end{tabular}%
  }
  \vspace{0pt}
\caption{\textbf{Concurrent Architectures on Cifar-10} -- Overall sampled metrics.}
\label{tb:cifar}
\vspace{-12pt}
\end{table}
\renewcommand{\arraystretch}{1.0}

%% file: tables/flower.tex
\renewcommand{\arraystretch}{1.0}
\begin{table}[t]
  \vspace{-5pt}
  \centering
  \small
  \resizebox{0.75\columnwidth}{!}{%
  \begin{tabular}{lcccc}
  
    \hline
    
    \multirow{2}{*}{}
    & \multirow{2}{0.12\columnwidth}{\centering\textbf{Mean Acc.}}
    & \multirow{2}{0.12\columnwidth}{\centering\textbf{Best Acc.}}
    & \multirow{2}{0.20\columnwidth}{\centering\textbf{Mean Acc./Param.}}
    & \multirow{2}{0.20\columnwidth}{\centering\textbf{Best Acc./Param.}} \\ \\
    
    \hline
    
    \textbf{ResNet-50}
    & 87.80
    & 87.80
    & 3.43
    & 3.43 \\
    
    \textbf{ER}
    & 84.88
    & 86.20
    & 2.11
    & 2.43 \\
    
    \textbf{BA}
    & 82.91
    & 84.62
    & 2.41
    & 2.91 \\
    
    \textbf{WS}
    & 81.46
    & 86.57
    & 3.17
    & 3.10 \\
    
    \textbf{DP}
    & 84.66
    & 86.69
    & 3.19
    & 3.28 \\

  \end{tabular}%
  }
  \vspace{-0pt}
\caption{\textbf{Concurrent Architects on Flower-102} -- Overall sampled metrics.}
\vspace{-20pt}
\label{tb:flower}
\end{table}
\renewcommand{\arraystretch}{1.0}

%% file: tex/conclusion.tex
In this work, we proposed concurrent architectures that break the single-chain of dependencies, a common bias in modern architecture designs. We showed that these architectures are concurrent and have more distribution opportunities for reducing the inference time while achieving competitive accuracy. Since we discover that previous NAS studies were implicitly biased in creating a sequential model, we introduced a new generator that naturally creates concurrent architectures. To quantitatively compare concurrent architectures, we proposed the concurrency score that encapsulates critical metrics in distribution.

%% file: tex/appendix.tex
\begin{figure}[!h]
  \vspace{-20pt}
  \centering
  \includegraphics[width=0.94\linewidth]{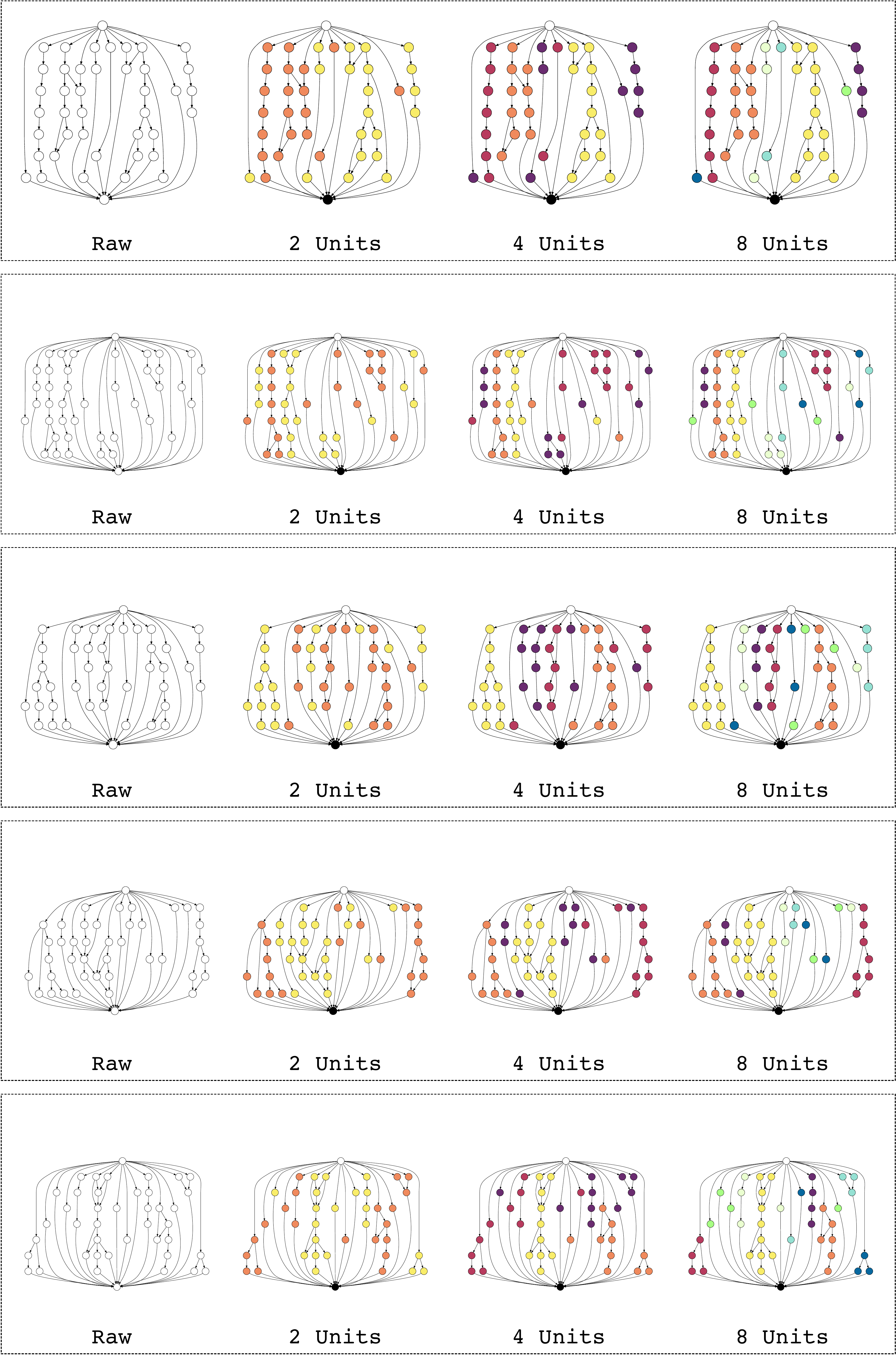}
  \vspace{-10pt}
  \caption{\textbf{Random Neural Network Distribution} -- This gives 5 examples of raw random generated neural networks, their distributions on two, four and eight units.}
  \label{fig:apdix:net-dist}
\end{figure}

\subsection{Distribution}
To distribute the generated networks according to the number of units, we first group node in the same sequential path together to minimize the communication overhead. The detailed algorithm of grouping can be found in \cref{apdix:grouping}. After the nodes in the graph are grouped together, we use heuristic-based greedy algorithm~\cref{apdix:lb} to distribute all nodes to units. The objective of the algorithm is to balance the workload. To make the load balancing simple, we assume the final goal is that each unit performs a similar amount of computations. Ultimately, this process can be improved using various other techniques that currently is out of the scope of this paper. Here, we provide an example of our process, which starts from network generation to workload distribution.

\subsubsection{Network Generation}
~\cref{fig:apdix:net-dist} demonstrates a example of raw random neural network generated. This network is later fed into a grouping and distribution algorithm to decide which unit runs which nodes.

\subsubsection{Distribution to 2,4 and 8 Units}
~\cref{fig:apdix:net-dist} shows network distribution on 2,4 and 8 units. The coloring marks the node is distributed on which unit. Because all units need to run the computations of the first node, we leave it as a common node (this could be just a scatter operation). In addition, for the last node, an extra unit is needed to merge all results together, so we mark that unit as black (this could be just a gather operation).

\begin{figure}[!b]
  \vspace{-15pt}
  \centering
  \includegraphics[width=0.7\linewidth]{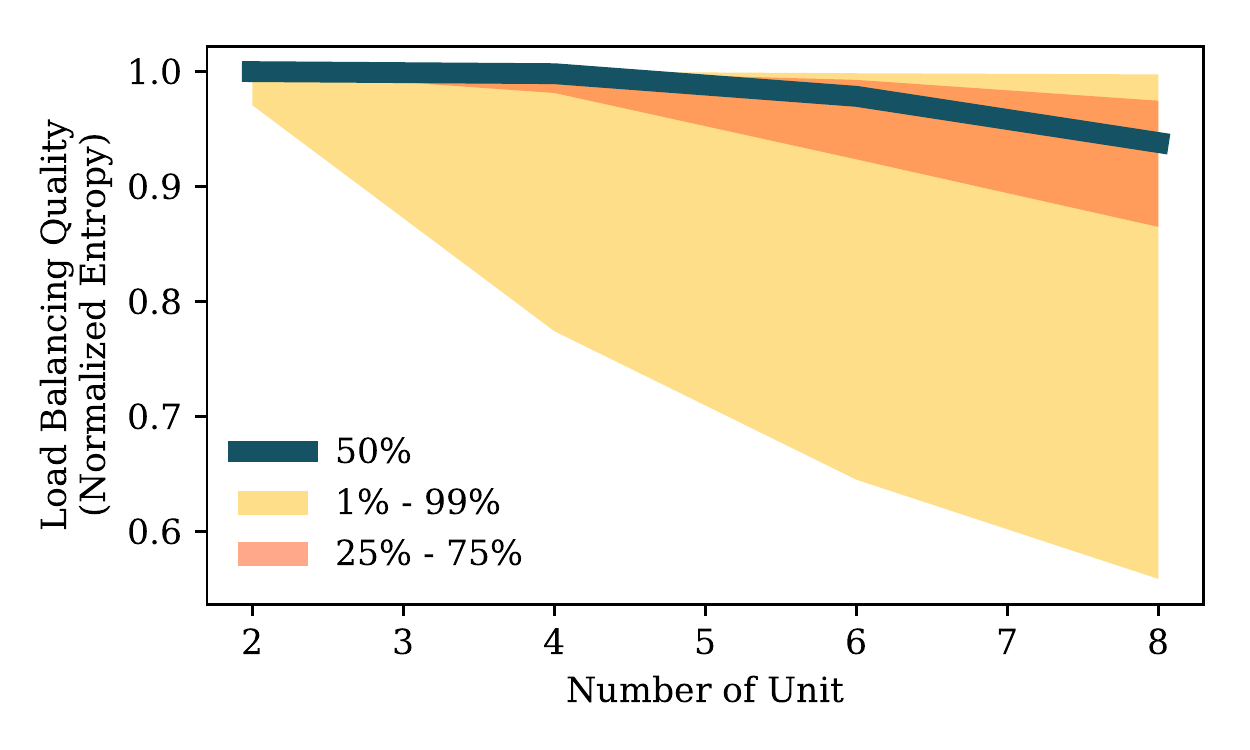}
  \vspace{-10pt}
  \caption{\textbf{Load Balance Quality} -- The load balance quality analysis on two, four, six and eight units compared to the normalized Shannon entropy value.}
  \label{fig:apdix:lb}
\end{figure}

\subsubsection{Load Balancing} 
From the graphs, we observe that the current grouping and distribution algorithm does well load balancing under the scenario with a small number of units. The quality of load balancing affects the final inference latency, because the final results may slow down due to a bottleneck node, which happens when unbalanced loads exist. We conduct a load balance quality study as well as shown in~\cref{fig:apdix:lb}. We use normalized Shannon entropy value to indicate the load balancing quality (the higher the number represents the load is more balanced, and $1$ means the load is perfectly balanced across distribution units). In the~\cref{fig:apdix:lb}, we showcase the median, $25\% - 75\%$ percentile, and $1\% - 99\%$ percentile load balancing qualities. We observe that as the number of distribution units increases, the overall load balancing quality downgrades and the variation of quality increases. We aim to develop distribution algorithms with higher quality; however, currently, our aim in this paper is showing that parallel inference computations of a single request is a viable option and should be studied more.

\subsubsection{Performance Scaling}
\begin{figure}[!t]
  \vspace{-0pt}
  \centering
  \includegraphics[width=1.0\linewidth]{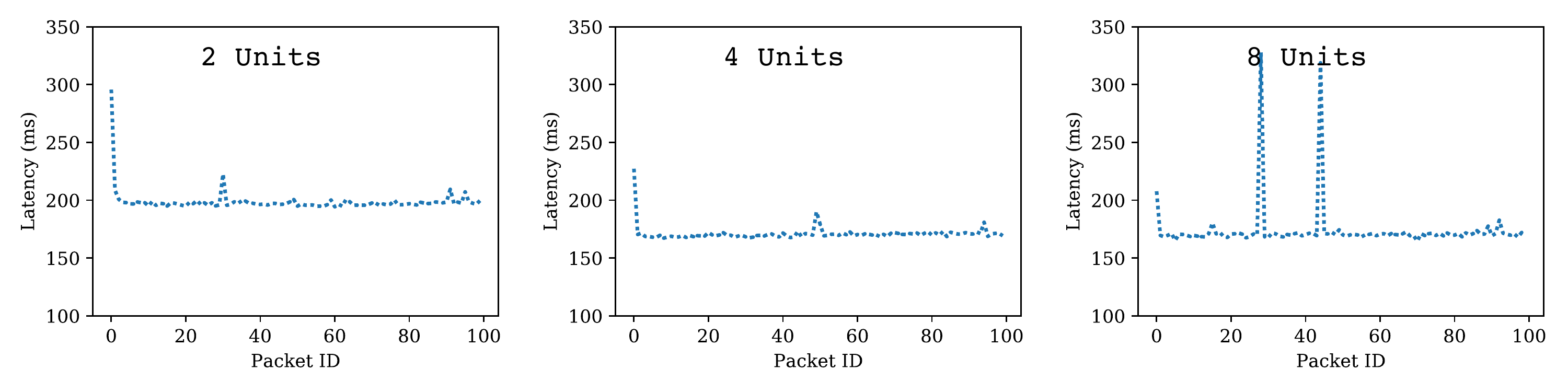}
  \vspace{-10pt}
  \caption{\textbf{Performance Scaling} -- the random neural network latency on two, four, and eight distribution units.}
  \vspace{-10pt}
  \label{fig:apdix:perf}
\end{figure}

As the final step, we also conduct a study on performance scaling. We use a total of $10$ AWS t2.micro EC2 instances for performance evaluation. Each instance is equipped with only $1$ vCPU and $1$ GB memory. The specification are chosen to emulate edge units with limited compute and memory that have a higher computational cost (remember that constants in the Equation~\ref{equ:cs} give higher priority to communication). As shown in~\cref{fig:apdix:perf}, the inference latency improves when the system has more distribution units. However, The latency stops to decrease as the number of distribution units becomes $8$, because the workload is not well balanced on each unit, as shown in our load balancing study. In this example, the bottleneck unit in the system causes longer latency for the entire system.

